\documentclass[preprints,article,accept,moreauthors,pdftex]{Definitions/mdpi} 

\firstpage{1} 
\pubvolume{xx}
\issuenum{1}
\articlenumber{5}
\pubyear{2020}
\copyrightyear{2020}
\history{Received: date; Accepted: date; Published: date}

\usepackage{subcaption}
\usepackage{adjustbox}
\usepackage{amssymb}

\usepackage{algorithm}
\usepackage[noend]{algpseudocode}
\usepackage{multirow, booktabs}
\usepackage{makecell}
\usepackage{framed}
\usepackage{stackengine}

\newlist{inlinelist}{enumerate*}{1}
\setlist*[inlinelist,1]{label=\roman*),itemjoin={{, }},itemjoin*={{, and }}}

\usetikzlibrary{arrows}
\usetikzlibrary{shapes}

\usepackage{pgfplotstable}
\pgfplotsset{compat=1.15}
\usepgfplotslibrary{statistics}
\usepackage{bm}

\def\eqref#1{equation~\ref{#1}}
\def\Eqref#1{Equation~\ref{#1}}

\def\1{\bm{1}}
\newcommand{\mathdataset}{\sD}
\newcommand{\trainset}{{\sD}_{\mathrm{train}}}

\def\vs{{\bm{s}}}

\def\vv{{\bm{v}}}

\def\vx{{\bm{x}}}
\def\vy{{\bm{y}}}

\def\mD{{\bm{D}}}

\def\mL{{\bm{L}}}

\def\mS{{\bm{S}}}

\def\mV{{\bm{V}}}

\def\mX{{\bm{X}}}

\DeclareMathAlphabet{\mathsfit}{\encodingdefault}{\sfdefault}{m}{sl}
\SetMathAlphabet{\mathsfit}{bold}{\encodingdefault}{\sfdefault}{bx}{n}

\def\gG{{\mathcal{G}}}

\def\adjmatrix{{\mathbf{\mathcal{A}}}}

\def\sD{{\mathbb{D}}}
\def\sE{{\mathbb{E}}}

\def\sS{{\mathbb{S}}}

\def\sV{{\mathbb{V}}}

\def\emAdjacency{{\mathcal{A}}}

\def\emD{{D}}

\newcommand{\R}{\mathbb{R}}

\DeclareMathOperator*{\argmax}{arg\,max}

\definecolor{imtatlantique}{rgb}{0.005,0.715,0.867}

\usepackage{xpatch}

\xpatchcmd{\algorithmic}{\itemsep\z@}{\itemsep=20pt}{}{}

\usetikzlibrary{decorations.markings}
\tikzset{
    set arrow inside/.code={\pgfqkeys{/tikz/arrow inside}{#1}},
    set arrow inside={end/.initial=>, opt/.initial=},
    /pgf/decoration/Mark/.style={
        mark/.expanded=at position #1 with
        {
            \noexpand\arrow[\pgfkeysvalueof{/tikz/arrow inside/opt}]{\pgfkeysvalueof{/tikz/arrow inside/end}}
        }
    },
    arrow inside/.style 2 args={
        set arrow inside={#1},
        postaction={
            decorate,decoration={
                markings,Mark/.list={#2}
            }
        }
    },
}

\theoremstyle{definition}

\Title{Representing Deep Neural Networks Latent Space Geometries with Graphs}

\Author{Carlos Lassance $^{1}$\orcidA{}, Vincent Gripon $^{1}$\orcidB{} and Antonio Ortega $^{2}$}

\AuthorNames{Carlos Lassance, Vincent Gripon and Antonio Ortega}

\address{%
$^{1}$ \quad IMT Atlantique; carlos.rosarkoslassance at imt-atlantique dot fr and vincent.gripon at imt-atlantique dot fr\\
$^{2}$ \quad University of Southern California; antonio dot ortega at ee dot usc dot edu}

\corres{Correspondence: carlos.rosarkoslassance at imt-atlantique dot fr}

\abstract{Deep Learning (DL) has attracted a lot of attention for its ability to reach state-of-the-art performance in many machine learning tasks. 
The core principle of DL methods consists in training composite architectures in an end-to-end fashion, where inputs are associated with outputs trained to optimize an objective function.
Because of their compositional nature, DL architectures naturally exhibit several intermediate representations of the inputs, which belong to so-called latent spaces.
When treated individually, these intermediate representations are most of the time unconstrained during the learning process, as it is unclear which properties should be favored.
However, when processing a batch of inputs concurrently, the corresponding set of intermediate representations exhibit relations (what we call a geometry) on which desired properties can be sought.
In this work, we show that it is possible to introduce constraints on these latent geometries to address various problems. In more details, we propose to represent geometries by constructing similarity graphs from the intermediate representations obtained when processing a batch of inputs. By constraining these Latent Geometry Graphs (LGGs), we address the three following problems: i) Reproducing the behavior of a teacher architecture is achieved by mimicking its geometry, ii) Designing efficient embeddings for classification is achieved by targeting specific geometries, and iii) Robustness to deviations on inputs is achieved via enforcing smooth variation of geometry between consecutive latent spaces.
Using standard vision benchmarks, we demonstrate the ability of the proposed geometry-based methods in solving the considered problems.}

\keyword{graph signal processing; deep learning; robustness; compression}

\begin{document}


\section{Introduction}

In recent years, Deep Learning (DL) methods have achieved state of the art performance in a vast range of machine learning tasks, including image classification~\citep{tan2019efficientnet} and multilingual automatic text translation~\citep{edunov2018understanding}. A DL architecture is built by assembling elementary operators called \emph{layers}~\citep{dlbook}, some of which contain trainable parameters. Due to their compositional nature, DL architectures exhibit intermediate representations when they process a given input. These intermediate representations lie in so-called latent spaces.

DL architectures are typically trained to minimize a loss function computed at their output. This is performed using a variant of the stochastic gradient descent algorithm that is backpropagated through the multiple layers to update the corresponding parameters. To accelerate the training procedure, it is very common to process batches of inputs concurrently. In such a case, a global criterion over the corresponding batch (e.g. the average loss) is backpropagated.

The training procedure of DL architectures is thus performed in an \emph{end-to-end} fashion. This end-to-end characteristic of DL refers to the fact that intermediate representations are unconstrained during training. This property has often been considered as an asset in the literature~\citep{lecun2015deep} which presents deep learning as a way to replace ``hand-crafted'' features by automatic differentiation. As a matter of fact, using these hand-crafted features as intermediate representations can cause sub-optimal solutions~\citep{lecun2018limits}. On the other hand, completely removing all constraints on the intermediate representations can cause the learning procedure to exhibit unwanted behavior, such as susceptibility to deviations of the inputs~\citep{cisse2017parseval,hendrycks2019robustness,lassance2020laplacian}, or redundant features~\citep{park2019rkd,lassance2020deep}. 

In this work we propose a new methodology aiming at enforcing desirable properties on intermediate representations. Since training is organized into batches, we achieve this goal by constraining what we call the \emph{latent geometry} of data points within a batch. This geometry refers to the relative position of data points within a specific batch, based on their representation in a given layer.
While there are many problems for which specific intermediate layer properties are beneficial, in this work we consider three examples. First, we explore compression via \emph{knowledge distillation} (KD)~\citep{hinton2014distillation,park2019rkd,koratana2019lit,lassance2020deep}, where the goal is to supervise the training procedure of a small DL architecture (called the student) with a larger one (called the teacher). Second, we study the design of efficient embeddings for classification~\citep{hermans2017defense,bontonou2019smoothness}, in which the aim is to train the DL architecture to be able to extract features that are useful for classification (and could be used by different classifier) rather than using classification accuracy as the sole performance metric. Finally, we develop techniques to increase the robustness of DL architectures to deviations of their inputs~\citep{cisse2017parseval,hendrycks2019robustness,lassance2020laplacian}.

To address the three above-mentioned problems, we introduce a common methodology that exploits the latent geometries of a DL architecture. More precisely, we propose to formalize latent geometries by defining similarity graphs. In these graphs, vertices are data points in a batch and an edge weight between two vertices is a function of the relative similarity between the corresponding intermediate representations at a given layer. We call such a graph a \textit{latent geometry graph (LGG)}. In this paper we show that  intermediate representations with desirable properties can be obtained by imposing constraints on their corresponding LGGs. 
In the context of KD, similarity between teacher and student is favored by minimizing the discrepancy between their respective LGGs. For efficient embedding designs, we propose a LGG-based objective function that favors disentanglement of the classes. Lastly, to improve robustness, we enforce smooth variations between LGGs corresponding to pairs of consecutive layers at any stage, from input to output, in a given architecture. Enforcing smooth variations between the LGGs of consecutive layers provides some protection against noisy inputs, since small changes in the input are less likely to lead to a sharp transition of the network's decision.


This paper is structured as follows, we first discuss related work in Section~\ref{related_work}. We then introduce the proposed methodology in Section~\ref{methodology}. Then we present the three applications, namely knowledge distillation, design of classification feature vectors and robustness improvements, in Section~\ref{applications}. Finally, we present a summary and a discussion on future work in Section~\ref{conclusion}.

\section{Related work}\label{related_work}


As previously mentioned, in this work we are interested in using graphs to ensure that latent spaces of DL architectures have some desirable properties. The various approaches we introduce in this paper are based on  our previous contributions~\citep{bontonou2019smoothness,lassance2020deep,lassance2020laplacian}. 
However, in this paper they are presented for the first time using a unified methodology and formalism. 
While we deployed these ideas already in a few applications, by presenting them in a unified form our goal is to provide a broader perspective of these tools, and to encourage their use for other problems. 

In what follows, we introduce related work found in the literature. We start by comparing our approach with others that also aim at enforcing properties on latent spaces. Then we discuss approaches that mix graphs and intermediate (or latent)  representations in DL architectures. Finally we discuss methods related to the applications highlighted in this work: \begin{inlinelist} \item knowledge distillation \item latent embeddings \item robustness \end{inlinelist}.

\textbf{Enforcing properties on latent spaces:} A core goal of our work is to enforce desirable properties on the latent spaces of DL architectures, more precisely \begin{inlinelist} \item consistency with a teacher network \item class disentangling \item smooth variation of geometries over the architecture  \end{inlinelist}. In the literature, one can find two types of approaches to enforce properties on latent spaces: \begin{inlinelist} \item directly designing specific modules or architectures~\citep{svoboda2019peernets,qian2019l2nonexpansive} \item modifying the training procedures~\citep{hinton2014distillation,hermans2017defense} \end{inlinelist}. The main advantage of the latter approaches is that one is able to draw from the vast literature in DL architecture design~\citep{he2016deep,krizhevsky2012imagenet} and use an existing architecture instead of having to design a new one. 

Our proposed unified methodology 
can be seen as an example of the second type of approaches, 
with two main advantages over competing techniques. 
First, by using relational information between the examples, instead of treating each one separately, we extend the range of proposed solutions. For example, relational knowledge distillation methods can be applied to any pair of teacher-student networks~\citep{park2019rkd} as relational metrics are dependent on the number of examples and not on the dimension of individual layers (see more details in the next paragraphs). Second, by using graphs to represent the relational information, we are able to
to exploit the rich literature in  graph signal processing \citep{shuman2013emerging} and use it to reason about the properties we aim at enforcing on latent spaces. We discuss this in more detail in Section~\ref{appli:kd} and Section~\ref{appli:robustness}.

\textbf{Latent space graphs:} In the past few years, there has been a growing interest in proposing deep neural network layers able to process graph-based inputs, also known as graph neural networks. For example, works such as~\citep{kipf2016semi,vialatte2018convolution} show how one can use convolutions defined in graph domains to improve performance of DL methods dealing with graph signals as inputs. 
The proposed methodology differs from these works in that it does not require inputs to be defined on an explicit graph. The graphs we consider here (LGGs) are proxies to the latent data geometry of the intermediate representations. 
As such, the proposed method could be considered orthogonal to graph neural network approaches. 
 Some recent work can be viewed as following ideas similar to those  introduced in this paper, with  applications in areas such as knowledge distillation~\citep{liu2019knowledge,lee2019graph}, robustness~\citep{svoboda2019peernets}, interpretability~\citep{anirudh2017influential} and generalization~\citep{gripon2018insidelook}. Despite sharing a common methodology, these works are not explicitly linked. This can be explained by the fact they were introduced independently around the same time, and have  different aims. We provide more details about how they are  connected with our proposed methodology in the following paragraphs. 

\textbf{Knowledge distillation:} Knowledge distillation is a DL compression method, where the goal is to use the knowledge acquired on a pre-trained architecture, called teacher, to train a smaller one, called student. Initial works on knowledge distillation considered each input independently from the others, an approach known as Individual Knowledge Distillation (IKD)~\citep{hinton2014distillation,romero2015fitnets,koratana2019lit}. As such, the student architecture mimics the intermediate representations of the teacher for each input used for training. The main drawback of IKD lies in the fact that it forces intermediate representations of the student to be of the same dimensions of that of the teacher. To deploy IKD in broader contexts, authors have proposed to disregard some of these intermediate representations~\citep{koratana2019lit} or to perform some-kind of dimensionality reduction~\citep{romero2015fitnets}.

On the other hand, the method we propose in Section~\ref{appli:kd} is based on a recent paradigm named Relational Knowledge Distillation (RKD)~\citep{park2019rkd}, which differs from IKD as it focuses on the relationship between examples instead of their exact positions in latent spaces. RKD has the advantage of leading to dimension-agnostic methods, such as the one described in this work. By defining graphs, its main advantage lies in the fact relationships between elements are considered relatively to each other.

Concurrently, other authors~\citep{liu2019knowledge,lee2019graph} have proposed methods similar to the one we present here~\citep{lassance2020deep}. In  \citep{liu2019knowledge}, unlike in our approach, dimensionality reduction transformations are added to the intermediate representations, in an attempt to improve the knowledge distillation. In \citep{lee2019graph}, LGGs are built using attention (similar to~\citep{velivckovic2017graph}). Among other differences, we show in Section~\ref{appli:kd} that constructing graphs that only connect data points from distinct classes can significantly improve accuracy.

\textbf{Latent embeddings:} In the context of classification, the most common DL setting is to train the architecture end-to-end with an objective function that directly generates a decision at the output. Instead, it can be beneficial to output representations well suited to be processed by a simple classifier (e.g. logistic regression). This framework is called feature extraction or latent embeddings, as the goal is to generate representations that are easy to classify, but without directly enforcing the way they should be used for classification. Such a framework is very interesting if the DL architecture is not going to be used solely for classification, but also for related tasks such as person re-identification~\citep{hermans2017defense}, transfer learning~\citep{hu2020exploiting} and multi-task learning~\citep{ruder2017overview}.

Many authors have proposed ways to train deep feature extractors. One influential example is~\citep{hermans2017defense}, where the authors use triplets to perform Deep Metric Learning. In each triplet, the first element is the example to train, the second is a positive example (e.g. same class) and the last is a negative one (e.g. different class). The aim is to result in triplets where the first element is closer to the second than to the last. 
In contrast, our method considers \textit{all} connections between examples of different classes, and can focus solely on separation (making all the negatives far) instead of clustering (making all the positives close), which we posit should lead to more robust embeddings in Section~\ref{appli:embeddings}.

Other solutions for generating latent embeddings propose alternatives to the classical $\argmax$ operator used to perform the decision at the output of a DL architecture. This can be done either by changing the output so that it is based on error correcting codes~\citep{dietterich1994solving} or is smoothed, either explicitly~\citep{szegedy2016rethinking} or by using the prior knowledge of another network~\citep{hinton2014distillation}. 

\textbf{Robustness of DL architectures:} In this work, we are interested in improving the robustness of DL architectures. We define robustness as the ability of the network to correctly classify inputs even if they are subject to small perturbations.  
These perturbations may be adversarial (designed exactly to force misclassification)~\citep{goodfellow2014adversarial} or incidental (due to external factors such as hardware defects or weather artifacts)~\citep{hendrycks2019robustness}. The method we present in Section~\ref{appli:robustness} is able to increase the robustness of the architecture in both cases. 
 Multiple works in the literature aim to improve the robustness of DL architectures following two main approaches: \begin{inlinelist} \item training set augmentation~\citep{madry2018towards} \item improved training procedure \end{inlinelist}. Our  contribution can be seen as an example of the latter approaches,  
but can be combined with augmentation-based methods, leading to an increase of performance compared to using the techniques separately~\citep{lassance2020laplacian}.

A similar idea was proposed in~\citep{svoboda2019peernets}, where the authors exploit graph convolutional layers in order to improve robustness of DL architectures applied to non-graph domains. 
 Their approach can be described as denoising the (test) input by using the training data. This differs from the method we propose in Section~\ref{appli:robustness}, which focuses on generating a smooth network function. As such, the proposed method is more general as it is less dependent on the training set.

\section{Methodology}\label{methodology}

In this section we first introduce basic  concepts from deep learning and graph signal processing  (Sections~\ref{methodo:dl} and~\ref{methodo:gsp}) and then  our proposed methodology (Section~\ref{methodo:proposed}). 

\subsection{Deep learning}\label{methodo:dl}

\begin{figure}[ht!]
  \begin{center}
    %
  \begin{tikzpicture}[thick,scale=1.0]    
  \node (00)[draw,diamond, rotate=270] at (-1,0) {Input: $\vx $};
  \node(0)[draw,rectangle,rotate=270] at (1,0) {Embedding layer};
  \node(1)[draw,rectangle, red,rotate=270] at (2,0) {Block 1};
  \node(2)[draw,rectangle, red,rotate=270] at (3,0) {Block 2};
  \node(3)[draw,rectangle, imtatlantique, rotate=270] at (4,0) {Block 3};
  \node(4)[draw,rectangle, imtatlantique, rotate=270] at (5,0) {Block 4};
  \node(5)[draw,rectangle, green!50!black, rotate=270] at (6,0) {Block 5};
  \node(6)[draw,rectangle, green!50!black, rotate=270] at (7,0) {Block 6};
  \node(7)[draw,rectangle, magenta, rotate=270] at (8,0) {Block 7};
  \node(8)[draw,rectangle, magenta, rotate=270] at (9,0) {Block 8};
  \node(9)[draw,rectangle, cyan, rotate=270] at (10,0) {Global pooling};
  \node(10)[draw,rectangle, green!25!black, rotate=270] at (11,0) {Classification layer};
  \node(11)[draw,ellipse, rotate=270] at (12,0) {Output: $\hat{\vy} = f(\vx)$};
  \path[->, >=stealth']
  (00) edge (0)
  (0) edge (1)
  (1) edge (2)
  (2) edge (3)
  (3) edge (4)
  (4) edge (5)
  (5) edge (6)
  (6) edge (7)
  (7) edge (8)
  (8) edge (9)
  (9) edge (10)
  (10) edge (11)
  ;
\end{tikzpicture}%

    \caption{Simplified depiction of a Resnet with eight residual blocks}
    \label{fig:resnet_simple}
  \end{center}
\end{figure}
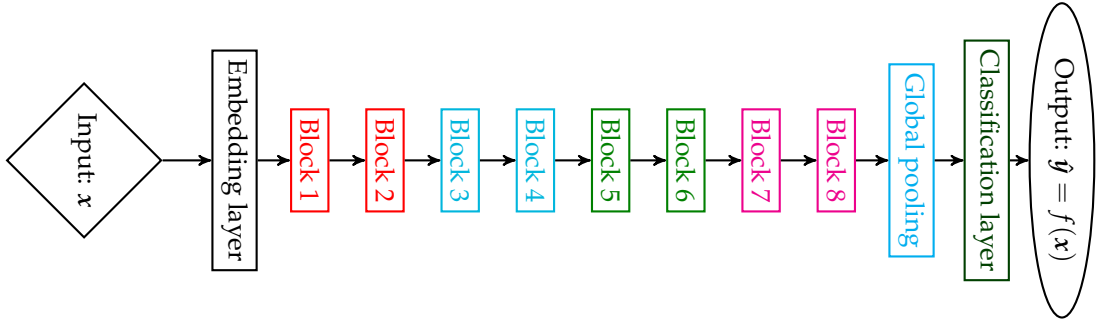

We start by introducing basic deep learning (DL) concepts, referring the reader to~\citep{dlbook} for a more in-depth overview. A DL architecture is an assembly of layers that can be mathematically described as a function $f$, often referred to as the ``network function'' in the literature, that associates an input tensor $\vx$ with an output tensor $\hat{\vy} = f(\vx)$. This function is characterized by a large number of trainable parameters $\theta$. 
In the literature, many different approaches have been proposed to assemble layers to obtain such network functions~\citep{he2016deep}. While layers are the basic unit, it is also common to describe architectures in terms of a series of \textit{blocks}, where a block is typically a small set of connected layers. This block representation allows us to encapsulate non-sequential behaviors, such as the residual connections of residual networks (Resnets)~\citep{he2016deep}, so that even though layers are connected in a more complex way, the blocks remain  sequential and the network function can be represented as a series of cascading operations:
\begin{equation}\label{compositional_equation}
    f = f^{\ell_\text{max}}\circ f^{\ell_\text{max}-1} \circ \dots \circ f^{\ell} \circ \dots \circ f^1, 
\end{equation}
where each function $f^{\ell}$ can represent a layer, or a block comprising several layers, depending on the underlying DL architecture. Thus, each block is associated with a subfunction $f^{\ell}$.
For example, in the context of Resnets~\citep{he2016deep}, the architecture is composed of blocks as depicted in Figure~\ref{fig:resnet_simple}.  

A very important concept for the remainder of this work is that of intermediate representations, which are the basis for the LGGs (defined in Section~\ref{methodo:proposed}) and corresponding applications (Section~\ref{applications}).

\begin{Definition}[Intermediate representation]\label{def_intermediate_representation}
We call intermediate representation of an input $\vx$ the output it generates at an intermediate layer or block. Starting from~\Eqref{compositional_equation}, and denoting $F^{\ell} = f^{\ell} \circ \dots \circ f^1,
$
we define the intermediate representation at depth $\ell$ for $\vx$ as $\vx^\ell \triangleq F^{\ell}(\vx)$. Or, said otherwise, $\vx^\ell$ is the representation of $\vx$ in the latent space at depth $\ell$.
\end{Definition}

Initially, the parameters $\theta$ of $f$ are typically drawn at random. They are then optimized during  the training phase so that $f$ achieves a desirable performance for the problem under consideration.
The dimension of the output of $f$ depends on the task. In the context of classification, it is common to design the network function such that the output has a dimension equal to the number of classes in the classification problem. In this case, for a given input, each coordinate of this final layer output is used as an estimate of the likelihood that the input belongs to the corresponding class. A network function correctly maps an input to its class if the output of the network function, $\hat{\vy}$, is close to the target vector of the correct class $\vy$.

\begin{Definition}[target vector]
  Each sample of the training set is associated with a target vector of dimension $C$, where $C$ is the total number of classes. Thus, the target vector of a sample of class $c$ is the binary vector containing 1 at coordinate $c$ and 0 at all other coordinates.
\end{Definition}

It is important to distinguish the target vector from the label indicator vector. The latter is defined on a batch of data points, instead of individually for each sample, as follows:

\begin{Definition}[label indicator vector]\label{label_indicator_vector}
  Consider a batch of $B$ data points. 
  The label indicator vector $\vv^c$ of class $c$ for this batch is the binary vector containing 1 at coordinate $i$ if and only if the $i$-th element of the batch is of class $c$ and 0 otherwise.
\end{Definition}

The purpose of a classification problem is to obtain a network function $f$ that outputs the correct class decision for any valid input $\vx$. In practice, it is often the case that the set of valid inputs $\sD$ is not finite, and yet we are only given a ``small'' number of pairs $(\vx,\vy)$, where $\vy$ is the output associated with $\vx$. The set of these pairs is called the dataset $\mathdataset$.
%
%
 During the training phase, the parameters are tuned using $\mathdataset$ and an objective function $\mathcal{L}$ that measures the discrepancy between the outputs of the network function and expected target indicator vectors, i.e., the discrepancy between $\hat{\vy} = f(\vx)$ and $\vy$. It is common to decompose the function $f$ into a feature extractor $\mathcal{F}$ and a classifier $\mathcal{C}$ as follows: $f = \mathcal{C}\circ\mathcal{F}$. In a classification task, the objective function is calculated over the outputs of the classifier and the gradients are backpropagated to generate a good feature extractor. Alternatively, to ensure that good latent embeddings are produced one can first optimize the feature extractor part of the architecture to optimize the features and then a classifier can be trained based on the resulting features  (which remain fixed or not)~\citep{hermans2017defense,bontonou2019smoothness}. We introduce an objective function designed for efficient latent embedding training in Section~\ref{appli:embeddings}.

Usually, the objective function is a loss function. It is minimized over a subset of the dataset that we call ``\textbf{training set}'' ($\trainset$). The reason to select a subset of $\mathdataset$ to train the DL architecture is that it is hard to predict the generalization ability of the trained function $f$. Generalization usually refers to the ability of $f$ to predict the correct output for inputs $\vx$ not in $\trainset$. A simple way to evaluate generalization consists in counting the proportion of elements in $\mathdataset - \trainset$ that are correctly classified using $f$. Obviously, this measure of generalization is not ideal, in the sense that it only checks generalization inside $\mathdataset$. This is why it is possible for a network that seems to generalize well to have trouble to classify inputs that are subject to deviations. In this case it is said that the DL architecture is not robust. We delve in more details on robustness in Section~\ref{appli:robustness}

In summary, a network function is initialized at random. Parameters are then tuned using a variant of the stochastic gradient descent algorithm on a dataset $\trainset$, and finally, training performance is evaluated on a validation set. Problematically, the best performance of deep learning architectures strongly depends on the total number of parameters they contain~\cite{hacene2019processing}. In particular it has been hypothesized that this dependence comes from the difficulty of finding a good gradient trajectory when the parameter space dimension is small~\cite{frankle2018the}. A common way to circumvent this problem is to rely on knowledge distillation, where a network with a large number of parameters is used to supervise the training of a smaller one. We introduce a graph-based method for knowledge distillation in Section~\ref{appli:kd}.

\subsection{Graph Signal Processing}\label{methodo:gsp}


As mentioned in the introduction, graphs are ubiquitous objects to represent relationships (called edges) between elements in a countable set (called vertices). In this section, we introduce the framework of Graph Signal Processing (\textbf{GSP}) which is central to our proposed methodology. Let us first formally define graphs:

\begin{Definition}[graph]
A graph $\gG$ is a tuple of sets $\langle \sV , \sE \rangle$, such that:  
\begin{enumerate}
 \item The finite set $\sV$ is composed of vertices $v_1, v_2, \dots $;
 \item The set $\sE$ is composed of pairs of vertices of the form ($v_i$,$v_j$) called edges.    
\end{enumerate}
\end{Definition}

It is common to represent the set $\sE$ using an edge-indicator symmetric adjacency matrix $\adjmatrix \in \R^{|\sV|\times |\sV|}$. Note that in this work we consider only undirected graphs corresponding to symmetric $\adjmatrix$ (i.e. $(v_i,v_j)\in \sE \Leftrightarrow (v_j,v_i) \in \sE$). 
In some cases, it is useful to consider (edge-)weighted graphs. In that case, the adjacency matrix can take values other than 0 or 1.

We can use $\adjmatrix$ to define the diagonal \textbf{degree matrix} $\mD$ of the graph as: 
\begin{equation}
    \emD_{i,j} = \left\{ \begin{array}{cl}\displaystyle{\sum_{j' \in \sV}{\emAdjacency_{i,j'}}} & \text{if } i = j\\ 0 & \text{otherwise}\end{array}\right.\;.
\end{equation} 

In the context of GSP, we consider not only graphs, but also graph signals. A graph signal is typically defined as a vector $\vs$. In this work we often consider a set of signals $\sS$ jointly. We group the signals in a matrix $\mS \in \R^{|\sV| \times |\sS|}$, where each of the columns is an individual graph signal $\vs$. An important notion in the remaining of this work is that of graph signal variation. 

\begin{Definition}[Graph signal variation]\label{variation}

The total variation $\sigma$ of a set of graph signals represented by $\mS$ 
is:
\begin{equation}
    \sigma = tr(\mS^\top \mL \mS),
\end{equation} 
where $\mL=\mD - \adjmatrix$ is the combinatorial Laplacian of the graph $\gG$ that supports $\mS$ and $tr$ is the trace function. We can also rewrite $\sigma$ as:
\begin{equation}\label{eq:sigma_rate}
    \sigma = tr(\mS^\top \mL \mS) = \sum_{i,j \in \sV}{\emAdjacency_{i,j} \sum_{\vs \in \sS} (\vs_i - \vs_j)^2} \;,
\end{equation} 
where $\vs_i$ represents the signal $\vs$ defined on vertex $v_i$. As such, the variation of a signal increases when vertices connected by edges with large weights have very different values.

\end{Definition}

\subsection{Proposed methodology}\label{methodo:proposed}

In this section we describe how to construct and exploit latent geometry graphs (LGGs) and illustrate the key ideas with a toy example. Given a batch $\mX$,  each LGG vertex corresponds to a sample in $\mX$ and each edge weight measures similarity between the corresponding data points. More specifically, LGGs are constructed as follows: 

\begin{enumerate}
    \item Generate a symmetric square matrix $\adjmatrix \in \R^{|\sV|\times |\sV|}$ using a similarity measure between  intermediate representations, at a given depth $\ell$, of data points in $\mX$. In this work we choose the cosine similarity when data is non-negative and an RBF similarity kernel based on the L2 distance otherwise;
    \item Threshold $\adjmatrix$ so that each vertex is connected only to its $k$-nearest neighbors;
    \item Symmetrize the resulting thresholded matrix: two vertices $i$ and $j$ are connected with edge weights $w_{ij}=w_{ji}$ as long one of the nodes was a $k$ nearest neighbor of the other. 
    \item (Optional) Normalize $\adjmatrix$ using its degree diagonal matrix $\mD$:  $\hat{\adjmatrix}= D^{-\frac{1}{2}} \adjmatrix D^{-\frac{1}{2}}$. 
\end{enumerate}

Given the LGG associated to some intermediate representation, we are able to quantify how well this  representation matches the  classification task under consideration by using the concept of label variation, a measure of graph signal variation for a signal formed as a concatenation of all label indicator vectors:

\begin{Definition}[Label variation]\label{label_variation}
Consider a similarity graph for a given batch $\mX$ (obtained from some intermediate layer), represented by an adjacency matrix $\adjmatrix$, and define a label indicator matrix $\mV$ obtained by concatenating label indicator vectors $\vv^c$ of each class. 
Label variation is defined as:
\begin{equation}
       \sigma = tr(\mV^{\top} \mL \mV)
       = \underbrace{\sum_{c}{\sum_{i,j\atop \vv^c_i \neq \vv^c_j}}}_{\text{sum over inputs of distinct classes}}{\adjmatrix_{i,j}\;.}
\end{equation}
\end{Definition}
If the graph is well suited for classification then most nodes will have immediate neighbors in the same class.  
Indeed, label variation is 0 if and only if data points that belong to distinct classes are not connected in the graph. Therefore smaller label variation is indicative of an easier classification task (well separated classes). 

\subsubsection{Toy example} 
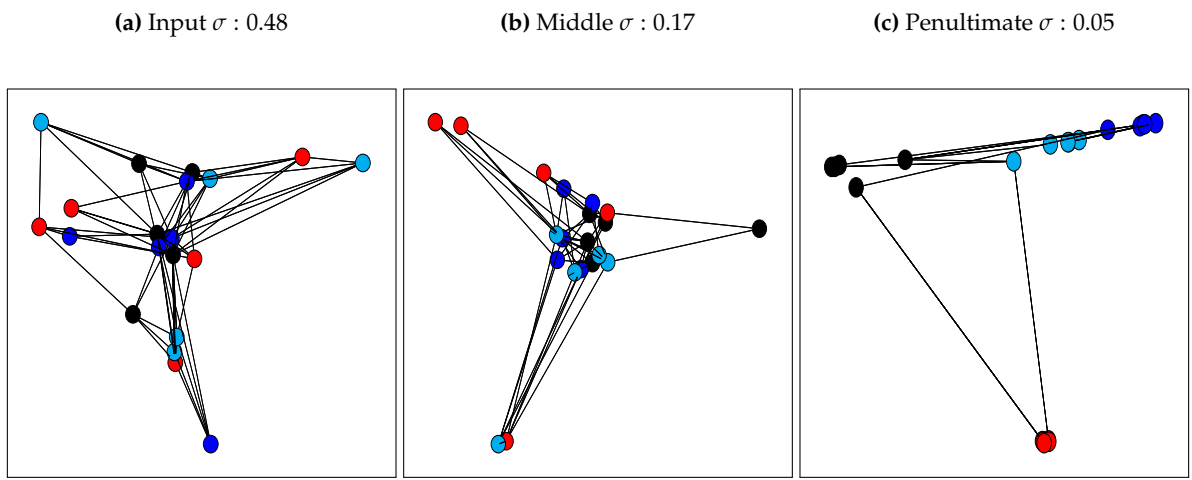
\begin{figure}[ht]
    \begin{center}
        \begin{subfigure}[ht]{0.32\linewidth}
            \centering
            \caption{Input $\sigma:0.48$}\label{figure-example-left}
            \vspace{0.5cm}
            \begin{framed}
                \begin{adjustbox}{width=\linewidth, height=\linewidth}
                    %
  \begin{tikzpicture}
\node(0)[draw,circle,fill=black,minimum size=1cm] at (-1.889310575076125,4.490010387064604) { };
\node(1)[draw,circle,fill=black,minimum size=1cm] at (0.4053975123372234,-0.6948448021855506) { };
\node(2)[draw,circle,fill=black,minimum size=1cm] at (1.706081316907556,3.988021866538311) { };
\node(3)[draw,circle,fill=black,minimum size=1cm] at (-0.6646649116525175,0.48212105700750774) { };
\node(4)[draw,circle,fill=black,minimum size=1cm] at (-2.297743427608364,-4.072312224776804) { };
\node(5)[draw,circle,fill=red,minimum size=1cm] at (-6.451350795794586,1.9539615014135254) { };
\node(6)[draw,circle,fill=red,minimum size=1cm] at (9.133965116591167,4.868347490973967) { };
\node(7)[draw,circle,fill=red,minimum size=1cm] at (0.5596491450531489,-6.818649568627745) { };
\node(8)[draw,circle,fill=red,minimum size=1cm] at (-8.623323303431091,0.897332968176517) { };
\node(9)[draw,circle,fill=red,minimum size=1cm] at (1.8488947350827993,-0.9203428882824205) { };
\node(10)[draw,circle,fill=blue,minimum size=1cm] at (-0.5511108354327212,-0.24178134873949297) { };
\node(11)[draw,circle,fill=blue,minimum size=1cm] at (2.949923711823374,-11.446186898442487) { };
\node(12)[draw,circle,fill=blue,minimum size=1cm] at (1.338309477149885,3.4758223667018395) { };
\node(13)[draw,circle,fill=blue,minimum size=1cm] at (-6.560938406690816,0.3594542679695189) { };
\node(14)[draw,circle,fill=blue,minimum size=1cm] at (0.3245743801801996,0.24728113313130612) { };
\node(15)[draw,circle,fill=cyan,minimum size=1cm] at (-8.498353718754732,6.8405140126127195) { };
\node(16)[draw,circle,fill=cyan,minimum size=1cm] at (0.5022543946512369,-6.207809887629006) { };
\node(17)[draw,circle,fill=cyan,minimum size=1cm] at (13.211855341470828,4.52719357520478) { };
\node(18)[draw,circle,fill=cyan,minimum size=1cm] at (2.8990742483641467,3.634785774314261) { };
\node(19)[draw,circle,fill=cyan,minimum size=1cm] at (0.6568165948293863,-5.362918782425374) { };
\path[-]
(0) edge (10)
(0) edge (12)
(0) edge (14)
(0) edge (15)
(0) edge (18)
(1) edge (5)
(1) edge (6)
(1) edge (7)
(1) edge (8)
(1) edge (9)
(1) edge (10)
(1) edge (11)
(1) edge (12)
(1) edge (13)
(1) edge (14)
(1) edge (16)
(1) edge (17)
(1) edge (18)
(1) edge (19)
(2) edge (6)
(2) edge (10)
(2) edge (12)
(2) edge (14)
(2) edge (15)
(2) edge (18)
(3) edge (5)
(3) edge (9)
(3) edge (10)
(3) edge (11)
(3) edge (12)
(3) edge (13)
(3) edge (14)
(3) edge (15)
(3) edge (16)
(3) edge (17)
(4) edge (7)
(4) edge (8)
(4) edge (10)
(4) edge (14)
(4) edge (16)
(4) edge (19)
(5) edge (1)
(5) edge (3)
(5) edge (12)
(5) edge (14)
(6) edge (1)
(6) edge (2)
(6) edge (12)
(6) edge (14)
(6) edge (17)
(6) edge (18)
(7) edge (1)
(7) edge (4)
(7) edge (10)
(7) edge (11)
(7) edge (14)
(7) edge (19)
(8) edge (1)
(8) edge (4)
(8) edge (13)
(8) edge (14)
(8) edge (15)
(9) edge (1)
(9) edge (3)
(9) edge (12)
(9) edge (14)
(9) edge (19)
(10) edge (0)
(10) edge (1)
(10) edge (2)
(10) edge (3)
(10) edge (4)
(10) edge (7)
(10) edge (18)
(10) edge (19)
(11) edge (1)
(11) edge (3)
(11) edge (7)
(11) edge (16)
(11) edge (19)
(12) edge (0)
(12) edge (1)
(12) edge (2)
(12) edge (3)
(12) edge (5)
(12) edge (6)
(12) edge (9)
(12) edge (15)
(12) edge (17)
(12) edge (18)
(13) edge (1)
(13) edge (3)
(13) edge (8)
(14) edge (0)
(14) edge (1)
(14) edge (2)
(14) edge (3)
(14) edge (4)
(14) edge (5)
(14) edge (6)
(14) edge (7)
(14) edge (8)
(14) edge (9)
(14) edge (16)
(14) edge (17)
(14) edge (18)
(14) edge (19)
(15) edge (0)
(15) edge (2)
(15) edge (3)
(15) edge (8)
(15) edge (12)
(16) edge (1)
(16) edge (3)
(16) edge (4)
(16) edge (11)
(16) edge (14)
(17) edge (1)
(17) edge (3)
(17) edge (6)
(17) edge (12)
(17) edge (14)
(18) edge (0)
(18) edge (1)
(18) edge (2)
(18) edge (6)
(18) edge (10)
(18) edge (12)
(18) edge (14)
(19) edge (1)
(19) edge (4)
(19) edge (7)
(19) edge (9)
(19) edge (10)
(19) edge (11)
(19) edge (14)
;
\end{tikzpicture}%

                \end{adjustbox}
            \end{framed}
        \end{subfigure}
        \begin{subfigure}[ht]{0.32\linewidth}
            \centering
            \caption{Middle $\sigma:0.17$}
            \vspace{0.5cm}
            \begin{framed}
                \begin{adjustbox}{width=\linewidth, height=\linewidth}
                    %
  \begin{tikzpicture}
\node(0)[draw,circle,fill=black,minimum size=1cm] at (1.6264377920550719,-1.1158676015958702) { };
\node(1)[draw,circle,fill=black,minimum size=1cm] at (1.2887021327745334,0.11041119335994165) { };
\node(2)[draw,circle,fill=black,minimum size=1cm] at (2.523970330368085,1.2077465949699382) { };
\node(3)[draw,circle,fill=black,minimum size=1cm] at (13.211855341470828,0.8507925707781421) { };
\node(4)[draw,circle,fill=black,minimum size=1cm] at (1.4265409883287428,1.6977845906874887) { };
\node(5)[draw,circle,fill=red,minimum size=1cm] at (2.6590076653311012,1.7613449187034016) { };
\node(6)[draw,circle,fill=red,minimum size=1cm] at (-4.370137919421716,-11.278598955084462) { };
\node(7)[draw,circle,fill=red,minimum size=1cm] at (-7.5039433862907154,6.725800949680593) { };
\node(8)[draw,circle,fill=red,minimum size=1cm] at (-1.7603889378184212,4.038491295413598) { };
\node(9)[draw,circle,fill=red,minimum size=1cm] at (-9.284995325252455,6.930012797382884) { };
\node(10)[draw,circle,fill=blue,minimum size=1cm] at (1.6268143736906127,2.3148676969135753) { };
\node(11)[draw,circle,fill=blue,minimum size=1cm] at (-0.8161878756544627,-0.9258881232660903) { };
\node(12)[draw,circle,fill=blue,minimum size=1cm] at (0.823964194486852,-1.4569690619861) { };
\node(13)[draw,circle,fill=blue,minimum size=1cm] at (-0.36679992724321386,3.134469092084732) { };
\node(14)[draw,circle,fill=blue,minimum size=1cm] at (-0.4762772858914145,0.3090104657277497) { };
\node(15)[draw,circle,fill=cyan,minimum size=1cm] at (-0.8684678672922943,0.5080672942093071) { };
\node(16)[draw,circle,fill=cyan,minimum size=1cm] at (2.6840157456620615,-1.0662131225753924) { };
\node(17)[draw,circle,fill=cyan,minimum size=1cm] at (-4.912030609201056,-11.446186898442487) { };
\node(18)[draw,circle,fill=cyan,minimum size=1cm] at (0.4032474751798473,-1.6422966897613056) { };
\node(19)[draw,circle,fill=cyan,minimum size=1cm] at (2.0846730947180228,-0.6567790071996354) { };
\path[-]
(0) edge (11)
(0) edge (12)
(0) edge (15)
(0) edge (17)
(1) edge (5)
(1) edge (9)
(1) edge (11)
(1) edge (12)
(1) edge (14)
(1) edge (15)
(1) edge (16)
(1) edge (18)
(2) edge (10)
(2) edge (12)
(2) edge (13)
(3) edge (5)
(3) edge (16)
(4) edge (5)
(4) edge (7)
(4) edge (8)
(4) edge (10)
(4) edge (11)
(4) edge (14)
(4) edge (16)
(5) edge (1)
(5) edge (3)
(5) edge (4)
(5) edge (12)
(5) edge (15)
(6) edge (12)
(6) edge (15)
(6) edge (16)
(6) edge (17)
(6) edge (18)
(7) edge (4)
(7) edge (14)
(7) edge (15)
(8) edge (4)
(8) edge (10)
(8) edge (13)
(8) edge (15)
(9) edge (1)
(9) edge (11)
(9) edge (14)
(10) edge (2)
(10) edge (4)
(10) edge (8)
(10) edge (19)
(11) edge (0)
(11) edge (1)
(11) edge (4)
(11) edge (9)
(11) edge (17)
(12) edge (0)
(12) edge (1)
(12) edge (2)
(12) edge (5)
(12) edge (6)
(12) edge (15)
(12) edge (16)
(12) edge (17)
(12) edge (18)
(12) edge (19)
(13) edge (2)
(13) edge (8)
(13) edge (15)
(13) edge (18)
(14) edge (1)
(14) edge (4)
(14) edge (7)
(14) edge (9)
(14) edge (15)
(14) edge (16)
(14) edge (17)
(14) edge (18)
(14) edge (19)
(15) edge (0)
(15) edge (1)
(15) edge (5)
(15) edge (6)
(15) edge (7)
(15) edge (8)
(15) edge (12)
(15) edge (13)
(15) edge (14)
(16) edge (1)
(16) edge (3)
(16) edge (4)
(16) edge (6)
(16) edge (12)
(16) edge (14)
(17) edge (0)
(17) edge (6)
(17) edge (11)
(17) edge (12)
(17) edge (14)
(18) edge (1)
(18) edge (6)
(18) edge (12)
(18) edge (13)
(18) edge (14)
(19) edge (10)
(19) edge (12)
(19) edge (14)
;
\end{tikzpicture}%

                \end{adjustbox}
            \end{framed}
        \end{subfigure}
        \begin{subfigure}[ht]{0.32\linewidth}
            \centering
            \caption{Penultimate $\sigma:0.05$}
            \vspace{0.5cm}
            \begin{framed}
                \begin{adjustbox}{width=\linewidth,height=\linewidth}
                    %
  \begin{tikzpicture}
\node(0)[draw,circle,fill=black,minimum size=1cm] at (-11.525384508447422,1.7101396676801712) { };
\node(1)[draw,circle,fill=black,minimum size=1cm] at (-13.211855341470828,2.7619806595924214) { };
\node(2)[draw,circle,fill=black,minimum size=1cm] at (-12.68885798075045,2.863728933448443) { };
\node(3)[draw,circle,fill=black,minimum size=1cm] at (-13.004210384784356,2.7660799232796918) { };
\node(4)[draw,circle,fill=black,minimum size=1cm] at (-8.148212546726922,3.1277178602260247) { };
\node(5)[draw,circle,fill=red,minimum size=1cm] at (1.3215882483136188,-11.338742222756022) { };
\node(6)[draw,circle,fill=red,minimum size=1cm] at (1.744910703595128,-11.302129277096794) { };
\node(7)[draw,circle,fill=red,minimum size=1cm] at (1.3703791527750167,-11.381636898683096) { };
\node(8)[draw,circle,fill=red,minimum size=1cm] at (1.7591182010079651,-11.375666017315055) { };
\node(9)[draw,circle,fill=red,minimum size=1cm] at (1.4438957590133634,-11.446186898442487) { };
\node(10)[draw,circle,fill=blue,minimum size=1cm] at (8.252234489652082,4.9458299031954835) { };
\node(11)[draw,circle,fill=blue,minimum size=1cm] at (9.1335016645675,5.010706617882217) { };
\node(12)[draw,circle,fill=blue,minimum size=1cm] at (8.055135753989079,4.853988687627749) { };
\node(13)[draw,circle,fill=blue,minimum size=1cm] at (5.834102104234779,4.670154451890277) { };
\node(14)[draw,circle,fill=blue,minimum size=1cm] at (8.378400224758586,4.957988496906147) { };
\node(15)[draw,circle,fill=cyan,minimum size=1cm] at (1.8726968364375105,3.92045859052007) { };
\node(16)[draw,circle,fill=cyan,minimum size=1cm] at (3.131036128269164,4.037076102011909) { };
\node(17)[draw,circle,fill=cyan,minimum size=1cm] at (-0.6536691308064769,3.0465380493952203) { };
\node(18)[draw,circle,fill=cyan,minimum size=1cm] at (3.8534044054947656,4.142002020826974) { };
\node(19)[draw,circle,fill=cyan,minimum size=1cm] at (3.0817862208779934,4.029971349811026) { };
\path[-]
(0) edge (5)
(0) edge (7)
(0) edge (9)
(0) edge (13)
(1) edge (17)
(2) edge (13)
(3) edge (17)
(4) edge (10)
(4) edge (13)
(4) edge (14)
(4) edge (15)
(4) edge (17)
(5) edge (0)
(6) edge (17)
(7) edge (0)
(8) edge (17)
(9) edge (0)
(10) edge (4)
(11) edge (18)
(12) edge (16)
(12) edge (18)
(12) edge (19)
(13) edge (0)
(13) edge (2)
(13) edge (4)
(14) edge (4)
(15) edge (4)
(16) edge (12)
(17) edge (1)
(17) edge (3)
(17) edge (4)
(17) edge (6)
(17) edge (8)
(18) edge (11)
(18) edge (12)
(19) edge (12)
;
\end{tikzpicture}%

                \end{adjustbox}
            \end{framed}
        \end{subfigure}
    \end{center}
    \caption{Graph representation example of 20 examples from CIFAR-10, from the input space (left) to the penultimate layer of the network (right). The different vertex colors represent the classes of the data points. To help the visualization, we only depict the edges that are important for the variation measure (i.e. edges between elements of distinct classes). Note how there are many more edges at the input (a) and how the number of edges decrease as we go deeper in the architecture (b and c).}
    \label{fig-example}
\end{figure}

In this example we visualize the relation between the classification task and the geometries represented by the graphs. To do so, we construct three similarity graphs for a very small subset\footnote{20 images from 4 classes are used, i.e. 5 images per class.} of the CIFAR-10 $\trainset$, one defined on the image space (i.e., computing the similarity between the 3072 dimensions of the raw input images) and two using the latent space representations of an architecture trained on the dataset. Such representations come from an intermediate layer (32,768 dimensions) and the penultimate layer (512 dimensions). What we expect to see qualitatively is that the classes will be easier to separate as we go deeper in the considered architecture, which should be reflected by the label variation score: the penultimate layer should yield the smallest label variation. We depict this example in Figure~\ref{fig-example}. Note that data points are placed in the 2D space using Laplacian eigenmaps~\citep{belkin2003laplacian}. As expected, we can qualitatively see the difference in separation from the image space to the latent spaces. We are also able to measure quantitatively how difficult it is to separate the classes using the label variation, which is lowest for the penultimate layer. For more details on how this example was generated we refer the reader to the appendix.

\subsubsection{Dimensionality and LGGs}

A key asset of the proposed methodology is that the number of vertices in the graph is independent of the dimension of the intermediate representations it was built from.
As such, it is possible to compare graphs built from latent spaces with various dimensions, as illustrated in Figure~\ref{fig-example}. Being  agnostic to dimension will be a key ingredient in the applications described in the following section.
It is important to note that, while the number of vertices is independent of the dimension of intermediate representations,  edge weights are a function of a similarity in the considered latent space, which can have very different meanings depending on the underlying dimensions.

In the context of DL architectures, a common choice of similarity measure is that of cosine. Interestingly, cosine similarity is well defined only for nonnegative data (as typically processed by a ReLU function) and bounded between 0 and 1. When data can be negative, we use a Gaussian kernel applied to the Euclidean distance instead. The problem remains that cosine or Euclidean similarities suffers from the curse of dimensionality. In an effort to reduce the influence of dimension when comparing LGGs obtained from latent spaces with distinct dimensions, in our experiments we make use of graph normalization, as defined in step 4 of LGG construction..
A more in-depth analysis and understanding of the influence of dimension on graph construction is a promising direction for future work, as improving the graph construction could benefit all applications covered in this work.

\section{Applications}\label{applications}

We now show how LGGs can be used in three specific applications: \begin{inlinelist} \item knowledge distillation \item latent embeddings \item robustness\end{inlinelist}. Details on the dataset used  can be found in the appendix.

\subsection{Knowledge distillation}\label{appli:kd}

First, we consider the case of knowledge distillation (KD). The goal of KD is to use the knowledge acquired by a pre-trained DL architecture that we call teacher $T$ to train a second architecture called student $S$. KD is normally performed in compression scenarios where the goal is to obtain an architecture $S$ that is less computationally expensive than $T$ while maintaining good enough generalization. In order to do so, KD approaches aim at making both networks consistent in their decisions. Consistency is usually achieved by minimizing a measure of discrepancy between the networks intermediate and/or final representations.

More formally, we can define the objective function of the student networks trained with knowledge distillation as:
\begin{equation}
  \mathcal{L} = \mathcal{L}_\text{task} + \lambda_{\text{KD}} \cdot \mathcal{L}_\text{KD}\;,\label{distill_loss}
\end{equation}
where $\mathcal{L}_\text{task}$ is typically the same loss that was used to train the teacher (e.g., cross-entropy), $\mathcal{L}_\text{KD}$ is the distillation loss and $\lambda_{\text{KD}}$ is a scaling parameter to control the importance of the distillation with respect to that of the task.

Recall that Individual Knowledge Distillation (IKD) requires  intermediate representations of $T$ and $S$ to be of the same dimensions. In order to avoid this drawback, Relational Knowledge Distillation (RKD) has been recently  proposed~\citep{park2019rkd,liu2019knowledge,lee2019graph}. Indeed, the method we introduce in this section is inspired by~\cite{park2019rkd}, where the authors propose to compare the distance obtained between the intermediate representations of a pair of data points in the teacher with the corresponding distance for the student. The goal then becomes to minimize the variation between these two distances. Interestingly, distances can be compared even if the corresponding  intermediate representations do not have the same dimension. However, we point out that forcing (absolute) distances to be similar is not necessarily desirable. As a matter of fact, it would be sufficient to consider distances \emph{relatively} to other pairs of data points. For example: consider a case where in the teacher latent space the distance between points A and B is 0.5 and the distance between points A and C is 0.25. Instead of forcing the student to have the same distances as well (0.5 and 0.25) we could just ensure that the AC distance is half of AB distance.

In this section we introduce a method that focuses on relative distances. We do so using normalized LGGs. The framework we consider, that we named Graph Knowledge Distillation (GKD) in~\citep{lassance2020deep}, consists in reducing the discrepancy between LGGs constructed in $T$ and $S$.

\textbf{Proposed approach (GKD):} Let us consider a layer in the teacher architecture, and the corresponding one in the student architecture. Considering a batch of inputs, we propose to build the corresponding graphs $\gG_T$ and $\gG_S$ capturing their geometries as described in Section~\ref{methodo:proposed}.


During training, we propose to use the following loss in Equation~\ref{distill_loss}: 
\begin{equation}
    \mathcal{L}_{\text{GKD}} = \mathcal{L}_d(\gG_T,\gG_S)\;,
\end{equation}
where $\mathcal{L}_d$ is the Frobenius norm between the adjacency matrices. In practice, many such additive terms can be added, one per pairs of layers to match in teacher and student architectures. Let us point out that the dimensions of latent spaces in $T$ and $S$ are likely to be very different. As such, the LGGs are susceptible to be hard to compare directly. This is why we make use of graph normalization (as described in step 4 of LGG graph construction), where similarities are considered relatively to each other. Despite not being ideal, graph normalization allows us to obtain considerable gains in accuracy, as illustrated in the following experiments.

The GKD loss measures the discrepancy between the adjacency matrices of teacher and student LGGs. In this way the geometry of the intermediate representations of the student will be enforced to converge to that of the teacher (which is already fixed). Our intuition is that since the teacher network is expected to generalize well to the test, mimicking its latent geometry should allow for better generalization of the student network as well. Moreover, since we use normalized LGGs, the similarities are considered relative to each other (so that each vertex on the graph has the same ``connection strength''), contrary to initial works in RKD~\cite{park2019rkd}, where each distance is taken in its absolute value and thus one sample can eclipse all the others (e.g. being too far away from the others).

\textbf{Experiments:}
To illustrate the gains we can achieve using GKD, we ran the following experiment. Starting from a WideResNet28-1~\citep{zagoruyko2016wide} teacher architecture with many parameters, for which  an error rate of $7.27\%$ is achieved on CIFAR-10, we first train a student without KD, called baseline, containing roughly 4 times less parameters. The resulting error rate is $10.37\%$. We then compared RKD and GKD. Results  in Table~\ref{kd-errorrate} show that GKD doubles the gains of RKD over the baseline.

\begin{table}[ht]
\centering
\caption{Error rate comparison on CIFAR-10 for KD methods.}
\begin{tabular}{r|cc|c}
  \multicolumn{1}{c}{Method}                   & \multicolumn{1}{c}{Error} & \multicolumn{1}{c}{Gain} & \multicolumn{1}{c}{Relative size}   \\ \hline 
  Teacher                  & 7.27\% & ---     & 100\% \\ \hline
  Baseline \footnotesize{(student without KD)}                 & 10.34\% & ---   & 27\%  \\   
  \hline
  \hline
  RKD-D~\citep{park2019rkd}          & 10.05\% & 0.29\%    & 27\%  \\ \hline
  GKD \footnotesize{(Ours)}~\citep{lassance2020deep}                      & \textbf{9.71\%} & \textbf{0.63\%}      & 27\%  \\ \hline

\end{tabular}
\label{kd-errorrate}
\end{table}

More details and experiments can be found in~\cite{lassance2020deep}, where it is shown that the gains can be explained by the fact the GKD student presents decisions that are more consistent with the teacher than the RKD student. Also, other experiments in~\citep{lassance2020deep} suggest that simple modifications to graph construction (e.g. connecting only data points of distinct classes) can improve even further the gains reported in Table~\ref{kd-errorrate}.

\subsection{Latent embeddings}\label{appli:embeddings}

We now present an objective function that consists in minimizing the label variation on the output of the considered deep learning architecture. The goal of our objective function is to train the DL architecture to be a good feature extractor for classification, as the LGGs generated by the features will have a very small label variation. This idea was originally proposed in~\citep{bontonou2019smoothness}. 


\textbf{Methodology:} 
Let us consider the representations obtained at the output of a deep learning architecture. We build the corresponding LGG $\gG$ as described in Section~\ref{methodo:proposed}. Then we propose to use the label variation on this LGG as the objective function to train the network. 
By definition, minimizing the label variation leads to maximizing the distances between outputs of different classes. Compared to the classic cross entropy loss, we observe that label variation as an objective function does not suffer from the same drawbacks, notably: the proposed criterion does not need to force the output dimension to match the number of classes, it can result in distinct clusters in the output domain for a same class (as it only deals with distances between examples from different classes), and it can leverage the initial distribution of representations at the output of the network function.

\textbf{Experiments:} To evaluate the performance of label variation as an objective function, we perform experiments with the CIFAR-10 dataset~\citep{krizhevsky2009learning} and using ResNet-18~\citep{he2016deep} as our DL architecture. 
%
%
In Table~\ref{variation-table} we report the performance of the deep architectures trained with the proposed loss compared with cross-entropy. We also report the relative Mean Corruption Error (MCE), which is a standard measure of robustness towards corruptions of the inputs over the CIFAR-10 corruption benchmark~\citep{hendrycks2019robustness}, where smaller values of MCE are better. We observe that label variation is a viable alternative to cross-entropy in terms of raw test accuracy, and that it leads to significantly better robustness. More details and experiments can be found in~\citep{bontonou2019smoothness}, where in particular we show how the initial distribution of data points is preserved throughout the learning process.

  \begin{table}[ht]
    \begin{center}
    \caption{Comparison between the cross-entropy and label variation functions.}    \label{variation-table}
    \begin{tabular}{r|c|c}
    \multicolumn{1}{c}{Cost function}               & \multicolumn{1}{c}{Clean test error} & \multicolumn{1}{c}{relative MCE} \\         \hline
    Cross-entropy               & {\bf 5.06}\%           & 100          \\
    Label Variation (ours)~\citep{bontonou2019smoothness}      & 5.63\%           & {\bf 90.33}  \\ \hline     
    \end{tabular}
    \end{center}
  \end{table}



\subsection{Improving DL robustness}\label{appli:robustness}


In this section we propose to use label variation as a regularizer applied at each layer of the considered architecture during training. We initially introduced this idea in~\citep{lassance2020deep}. As it is not desirable to enforce a small label variation at early layers in the architecture, the core idea is to ensure a smooth evolution on label variation from an intermediate representation to the next one in the processing flow.


Recall that networks are typically trained with the objective of yielding zero error for the training set. If error on the training set is (approximately) zero then any two examples with different labels can be separated by the network, even if these examples are very close to each other in the original domain. This means that the network function can create significant deformations of the space (i.e., small distances in the original domain map to larger distances in the final layers) and explains how an adversarial attack with small changes to the input can lead to changing the output decision given by the network. When we enforce smooth evolution of label smoothness, we precisely prevent such sudden deformations of space.

\textbf{Methodology:} Formally, denote $\ell$ the depth of an intermediate representation in the architecture. Let us consider a batch of inputs, and let us build the corresponding LGG $\gG_\ell$ as described in Section~\ref{methodo:proposed}. The proposed regularizer can be expressed as:

\begin{equation}\mathcal{L}_{\text{reg}} = \sum_\ell{|\sigma^{\ell+1} - \sigma^{\ell}|}\;,\label{eq-reg}\end{equation}
where $\sigma^\ell$ is the label variation on $\gG_\ell$. 
This proposed regularizer is then added to the objective function (loss) with a scaling hyperparameter $\gamma$.


\textbf{Experiments:} In order to stress the ability of the proposed regularizer in improving robustness. We consider a ResNet18 that we trained on CIFAR-10. We consider multiple settings. In a first one, we add adversarial noise to inputs (\citep{goodfellow2014adversarial}) and compare the obtained accuracy. In a second one, we consider agnostic corruptions (i.e. corruptions that do not depend on the network function) and report the relative MCE~\citep{hendrycks2019robustness}. Results are presented in Table~\ref{regularizer-results}. The proposed regularizer performs better than the raw baseline and existing alternatives in the literature~\cite{cisse2017parseval}. More details can be found in~\citep{lassance2020laplacian}.

\begin{table}[ht]
\begin{center}
\caption{Comparison of different methods on their clean error rate and robustness.}
\begin{tabular}{@{}r |cc | c @{}}
\multicolumn{1}{c}{Metric} & \multicolumn{2}{c}{Error rate} & \multicolumn{1}{c}{relative MCE}  \\  \hline
\multicolumn{1}{c}{Method} & \multicolumn{1}{c}{Clean} & \multicolumn{1}{c}{Adversarial Attack~\citep{goodfellow2014adversarial}} & \multicolumn{1}{c}{Corruptions~\citep{hendrycks2019robustness}} \\ \hline \hline
Baseline  & 11.1\% & 66.3\% & 100   \\
Parseval~\citep{cisse2017parseval}  & \textbf{10.3}\% & 55.0\%  & 104.4    \\
Label variation regularizer (ours)~\citep{lassance2020laplacian}   & 13.2\% & \textbf{49.5}\% &\textbf{97.6}  \\ \hline

\end{tabular}
\label{regularizer-results}
\end{center}
\end{table}

\section{Conclusion}\label{conclusion}

In this work, we have introduced a methodology to represent latent space geometries using similarity graphs (i.e. LGG). We demonstrated the interest of such a formalism for three different problems: \begin{inlinelist} \item knowledge distillation \item latent embeddings \item robustness.\end{inlinelist}
With the ubiquity of graphs in representing relations between data elements, and the growing literature on Graph Signal Processing, we believe that the proposed formalism could be applied to many more problems and domains, including predicting generalization, improving performance in data-thrifty settings, and helping understanding how decisions are taken in a deep learning architecture.



Note that the proposed methodologies use straightforward techniques to build LGGs, and thus could be enriched with more principled approaches~\citep{kalofolias2018large,shekkizhar2020graph}. Another area of interest would be to build upon~\citep{svoboda2019peernets} and see what improvements may arise from the use of graph convolutional networks in domains that are not typically supported by graphs.

\vspace{6pt} 


\authorcontributions{Initial ideas were proposed jointly by AO, CL and VG. Initial investigation of the ideas was performed by CL. CL performed all simulations, prepared the figures, wrote the first draft and participated in the edition process. VG and AO supervised the project, and worked on the editing process from the original draft to the final version. }

\funding{Carlos Lassance was partially funded by the Brittany region in France. }

\acknowledgments{Most of the experiments realized in this work were done using GPUs gifted by NVIDIA. We would also like to acknowledge: M. Bontonou, G. Boukli Hacene, J. Tang and B. Girault for the discussions during the development of the methods presented here.}

\conflictsofinterest{The authors declare no conflict of interest.} 







\appendix
\section{CIFAR-10 dataset}

CIFAR-10 is a tiny ($32 \times 32$ pixels) image dataset extracted from the 80 million tiny images dataset~\citep{torralba2008tinyimages}. 
The 10 after the dataset names specify the number of classes of the problem. CIFAR-10 is composed of 60,000 images, being 50,000 images on the training set (5,000 per class), and 10,000 images on the test set (1,000 per class for CIFAR-10). 

\section{Details on the creation of the illustrative example}

We first sample images from the training set of CIFAR-10. We sample five images per class from four distinct classes.
We then input these 20 images on a Resnet18 trained on CIFAR-10 and keep the intermediate representations from the output of block 4 (that we call middle) and the global pooling (that we call penultimate). We refer the reader to Figure~\ref{fig:resnet_simple} for a visual description of where these blocks are placed in the overall architecture.

With these representations in hand (images, middle and penultimate) we can now use the framework described in Section~\ref{methodo:proposed} to generate LGGs. We use a $k$ of 5 to ensure that each vertex will have at least one connection with a vertex from another class. Finally we normalize the label variation so that 1 is the highest value possible (all connections between data points of different classes is equal to 1) and 0 is the lowest one (no connections between elements of different classes).

\externalbibliography{yes}
\bibliography{references.bib}

\end{document}